\documentclass{article}

\PassOptionsToPackage{numbers,sort&compress}{natbib}
 \usepackage[main, final]{main}

\usepackage[utf8]{inputenc} 
\usepackage[T1]{fontenc}    
\usepackage{hyperref}       
\usepackage{url}            
\usepackage{booktabs}       
\usepackage{amsfonts}       
\usepackage{nicefrac}       
\usepackage{microtype}      
\usepackage{xcolor}         
\usepackage{amsmath,amssymb,bm,booktabs,multirow}
\usepackage{algorithmic}
\usepackage{algorithm}
\usepackage{array}
\usepackage[caption=false,font=footnotesize,labelfont=rm,textfont=rm]{subfig}
\usepackage{textcomp}
\usepackage{stfloats}
\usepackage{verbatim}
\usepackage{graphicx}
\usepackage{utfsym}
\usepackage{threeparttable}
\usepackage[usestackEOL]{stackengine}
\usepackage[numbers,sort&compress]{natbib} 
\usepackage{wrapfig}

\title{Scaling Learned Image Compression Models up to \\1 Billion}

\author{%
  Yuqi Li, Haotian Zhang, Li Li, Dong Liu\thanks{Corresponding author: Dong Liu.}, Feng Wu\\
  University of Science and Technology of China \\
  \small \texttt{\{lyq010303, zhanghaotian\}@mail.ustc.edu.cn, \{lil1, dongeliu, fengwu\}@ustc.edu.cn} \\
}

\begin{document}

\maketitle

\begin{abstract}
Recent advances in large language models (LLMs) highlight a strong connection between intelligence and compression. Learned image compression, a fundamental task in modern data compression, has made significant progress in recent years. However, current models remain limited in scale, restricting their representation capacity, and how scaling model size influences compression performance remains unexplored. In this work, we present a pioneering study on scaling up learned image compression models and revealing the performance trends through scaling laws. Using the recent state-of-the-art HPCM model as baseline, we scale model parameters from 68.5 millions to 1 billion and fit power-law relations between test loss and key scaling variables, including model size and optimal training compute. The results reveal a scaling trend, enabling extrapolation to larger scale models. Experimental results demonstrate that the scaled-up HPCM-1B model achieves state-of-the-art rate-distortion performance. We hope this work inspires future exploration of large-scale compression models and deeper investigations into the connection between compression and intelligence.
\end{abstract}

\section{Introduction}

The recent advancements in natural language understanding and generation by large language models (LLMs), such as GPT~\cite{achiam2023gpt}, Qwen~\cite{yang2025qwen3}, and DeepSeek~\cite{guo2025deepseek}, have not only revolutionized natural language processing but also raised fundamental questions about the nature of intelligence. 
Some studies suggest a close connection between intelligence and the ability to compress information\cite{hutter2006prize, huang2024compression, li2024understanding}.
According to information theory\cite{rdtheory}, optimal data compression requires assigning shorter codewords to frequent symbols and longer codewords to rare ones, minimizing the expected negative log-likelihood of the data.
This objective is mathematically equivalent to maximizing the log-likelihood in probabilistic modeling, which is the principle behind training LLMs.
Therefore, advances in compression can be viewed as advances in the ability to model, predict, and reason about the world, which are generally regarded as characteristics of intelligence.

This connection raises a question: Do data compression models have the potential to exhibit intelligent properties? Among the various domains of data compression, image compression plays a crucial role, serving as a key technology in signal processing and communication. Traditional image compression standards such as JPEG\cite{wallace1991jpeg}, JPEG2000\cite{skodras2001jpeg}, and BPG\cite{bellard2015bpg} have been widely adopted over time.
Recent years have demonstrated the great success of learned compression techniques\cite{tang2025neural, jia2025towards}, especially learned lossy image compression\cite{Feng_2025_CVPR,Lu_2025_CVPR,li2025hpcm}, where models optimize a rate-distortion trade-off to retain essential visual and semantic information. Since Balle \textit{et al.} \cite{balle2017end} presented the pioneering work in 2016, many works have contributed to the transform\cite{zhu2022transformer, Feng_2025_CVPR}, quantization\cite{zhang2023uniform, guo2021soft}, and entropy coding\cite{he2022elic, Lu_2025_CVPR}.
Recently, Li \textit{et al.}\cite{li2025hpcm} proposed a learned image compression framework with hierarchical progressive context modeling (HPCM), surpassing the advanced traditional coding standard VVC\cite{bross2021overview} by around 20\%.
However, these models are relatively small in scale due to coding complexity constraints, potentially limiting their representation capacity. Moreover, the relationship between model size and compression performance remains unexplored, leaving the question of whether large-scale image compression models could yield significant gains or even reveal intelligent behaviors.

\begin{figure}[!t]
  \centering
    \includegraphics[width=0.7\linewidth]{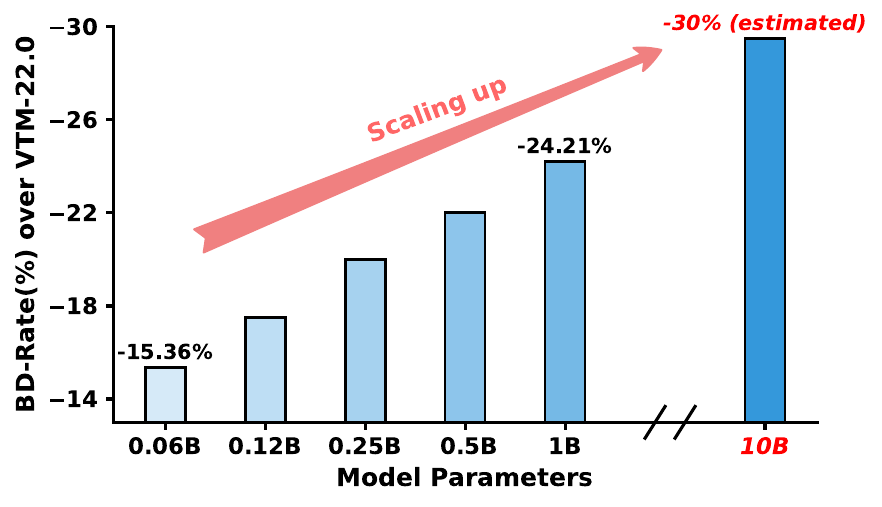}
    \caption{BD-Rate reduction over VTM-22.0 on Kodak dataset with different model parameter scales.
    The results demonstrate the performance improvement of our scaled-up HPCM models from 0.06B to 1B parameters, along with an extrapolated estimate for a 10B-parameter model based on scaling law analysis.
    }
    \label{fig1}
\end{figure}

To bridge this gap, we present a pioneering study on scaling up learned image compression models, offering a potential way to explore the link between large-scale compression models and intelligence. Specifically, building on the state-of-the-art HPCM\cite{li2025hpcm} framework, we scale the model parameters from 68.5 millions to 1 billion. As shown in Fig.~\ref{fig1}, with the power of increased model capacity, our HPCM-1B model can achieve superior compression performance.
Beyond performance improvements, we empirically investigate how compression performance changes with model size.
While such scaling behaviors, often referred to as scaling laws\cite{kaplan2020scaling}, have been extensively studied in LLMs\cite{hoffmann2022training} and vision foundation models\cite{oquab2023dinov2}, they remain unexplored in compression models. 
In this work, we conduct the first empirical validations on the scaling laws in learned compression models, revealing a predictable relationship between model size and compression performance.
This scaling behavior is similar to that of LLMs, suggesting the potential of large compression models as a tool for exploring the relationship between compression and intelligence. 
We hope our work inspires further research in large-scale compression models and deeper investigations into compression and intelligence.

\section{Related Work}

\subsection{Large Models and Scaling Laws}

Studies on large-scale models have revealed clear power-law scaling relationships: as model parameters, training data, and compute increase, task loss tends to decrease in a predictable manner. Kaplan \textit{et al.} \cite{kaplan2020scaling} first quantified this relationship for language models through cross-entropy scaling, while Hoffmann \textit{et al.} \cite{hoffmann2022training} refined it into the Chinchilla compute-optimal scaling rule, which balances model size and training tokens for maximum efficiency. Similar patterns have also been observed in multimodal generative models \cite{henighan2020scaling}, general deep learning tasks \cite{rosenfeld2021scaling}, and even image reconstruction \cite{klug2022scaling}, suggesting that predictable performance gains can be achieved when scaling is matched with sufficient data and compute.

In large language models (LLMs), representative models such as GPT-4 \cite{achiam2023gpt}, LLaMA 3 \cite{dubey2024llama}, Qwen3 \cite{yang2025qwen3}, and DeepSeek-R1 \cite{guo2025deepseek} demonstrate that scaling model capacity with high-quality data and stable training reliably enhances reasoning, generalization, and multilingual performance, aligning with compute-optimal scaling laws.

For vision foundation models (VFMs) and multimodal LLMs (MLLMs), works like DINOv2 \cite{oquab2023dinov2}, ViT-22B \cite{dehghani2023scaling}, and InternVL3 \cite{zhu2025internvl3} show that larger models pretrained on large, high-quality datasets deliver more transferable features and better zero/few-shot performance, with compute-optimal trends extending scaling benefits to vision and cross-modal tasks.

\subsection{Learned Image Compression}

Recent learned compression approaches \cite{li2024ustc,li2024uniformly,li2024object,li2024loop,tang2025neural,li2021deep,li2023neural,li2024neural,jia2025towards,hu2021fvc} have shown superior rate-distortion performance, typically following the joint optimization framework \cite{balle2017end} that integrates transform, quantization, and entropy model \cite{he2022elic,zou2022devil,liu2023learned,jiang2023mlic,li2023frequency,fu2024weconv,Feng_2025_CVPR,Lu_2025_CVPR,li2025hpcm}. 

Transform capacity has evolved from early convolutional designs \cite{balle2017end} to deeper residual and non-local attention structures \cite{chen2021end,cheng2020learned,zou2022devil,he2022elic}, invertible networks \cite{ma2020end}, transformer-based global modeling \cite{zhu2022transformer,liu2023learned,li2023frequency}, and linear attention variants \cite{Feng_2025_CVPR}. Entropy models have progressed from hyperprior \cite{balle2018variational} and autogressive context models \cite{minnen2018joint} to joint spatial-channel context exploitation \cite{he2021checkerboard,minnen2020channel,he2022elic,li2023neural}, multi-reference contexts \cite{jiang2023mlic}, dictionary-based references \cite{wu2025conditional, Lu_2025_CVPR}, and hierarchical progressive context modeling \cite{li2025hpcm}. Quantization and training strategies \cite{zhang2023uniform,guo2021soft,li2024deviation} have also been explored. Recently, Zhang \textit{et al.} \cite{zhang2025gap} analyze the gap between ideal and empirical rate-distortion function for lossy image compression, revealing the high potential of future lossy image compression technologies.
Besides, some studies have also leveraged LLMs for compression tasks~\cite{du2025large, chen2024large, li2024understanding}. 

However, despite these advances, none of the above studies investigate scaling up learned image compression models, leaving it unclear whether a large-scale learned image compression model could yield substantial gains or exhibit intelligent behaviors.

\begin{figure}[!t]
    \centering
    \centerline{\includegraphics[width=1.0\textwidth]{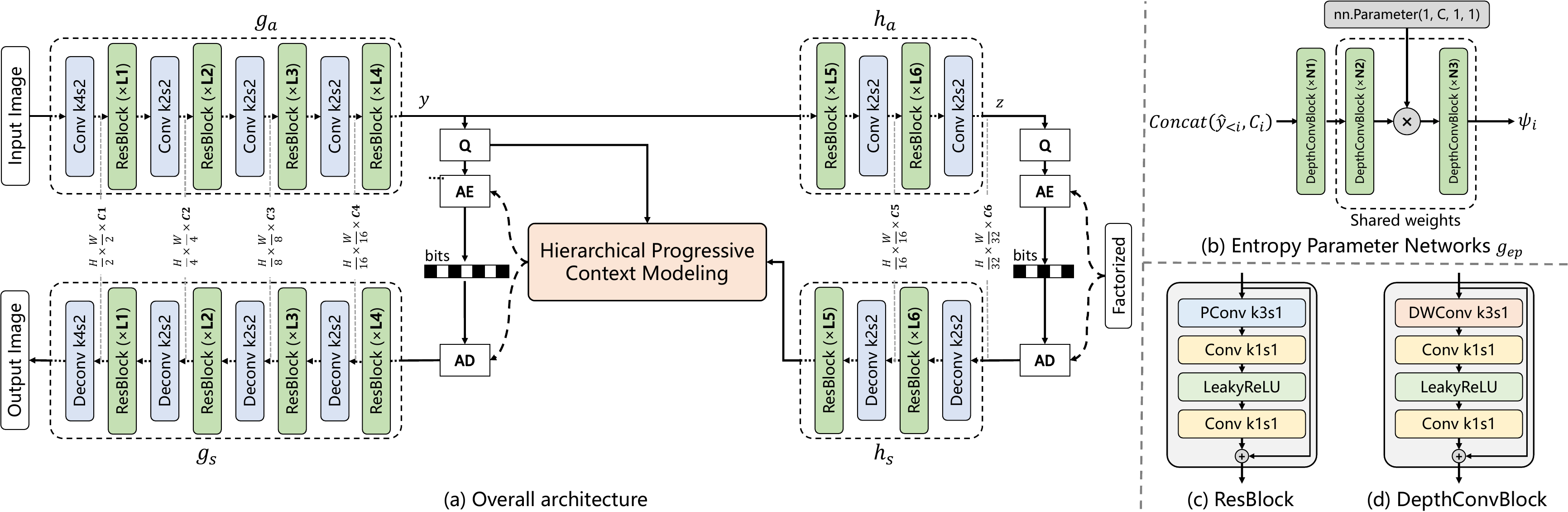}}
    \caption{Overall architecture of the proposed model, following the design of HPCM~\cite{li2025hpcm}. (a) Main architecture. `k2s2' denotes a convolution layer with kernel size as 2 and stride as 2. (b) Structure of the entropy parameter network $g_{ep}$. (c) and (d) Structures of the ResBlock and DepthConvBlock modules, respectively.
    }
    \label{arch}
\end{figure}

\section{Methods}

\subsection{Overview}
Our approach builds upon the HPCM framework\cite{li2025hpcm}. The overall architecture is shown in Fig.~\ref{arch}.
The encoder first applies an analysis transform $g_a$ to the input image $x \in \mathbb{R}^{3 \times H \times W}$, producing latent features
\begin{equation}
    y=g_a(x|\phi) \in \mathbb{R}^{C \times H \times W}
\end{equation}
The latents are quantized to $\hat{y} = Q(y)$, which are then losslessly compressed via entropy coding using a learned probability model $q_{\hat{Y}}(\hat{y})$. The decoder reconstructs $\hat{x}$ from $\hat{y}$ through a synthesis transform
\begin{equation}
    \hat{x}=g_s(\hat{y}|\theta) \in \mathbb{R}^{3 \times H \times W}
\end{equation}
where $\phi$ and $\theta$ denote the trainable parameters of the analysis and synthesis transforms, respectively.

Following the prior work \cite{zhang2024ggm}, we model the distribution of $\hat{y}$ as a generalized Gaussian model $\mathcal{N}_{\beta}(\mu,\alpha)$ with the shape parameter $\beta$ fixed as 1.5. The entropy model outputs the mean and scale parameters, which are estimated from the hyperprior module and the hierarchical progressive context model (HPCM). 
The hyperprior extracts side information $z$ through a hyper-analysis transform $z = h_a(y | \phi_h)$, which is quantized to $\hat{z} = Q(z)$. This side information is then decoded via a hyper-synthesis transform $h_s(\hat{z} | \theta_h)$ to provide initial entropy parameters. Here, $\phi_h$ and $\theta_h$ represent the trainable parameters of the hyper-analysis and hyper-synthesis transforms, respectively.
In HPCM, the latents are partitioned into multiple groups and coded sequentially. At the $i$-th coding step, the entropy parameters are refined using both the already-coded latents $\hat{y}_{<i}$ and the decoded side information:
\begin{equation}
\mu_{i}, \alpha_{i} = \operatorname{HPCM}(\hat{y}_{<i}, h_s(\hat{z} | \theta_h)).
\end{equation}
The bitrate of $\hat{y}$ is computed as
\begin{equation}
\begin{aligned}
    \mathcal{R}(\hat{y})&=\sum_{i} -\log_2 q_{\hat{Y}}(\hat{y}_i) \\
    q_{\hat{Y}}(\hat{y}_i)&=c(\frac{\hat{y}_i-\mu_i+0.5}{\alpha_i})-c(\frac{\hat{y}_i-\mu_i-0.5}{\alpha_i})
\end{aligned}
\end{equation}
where $c(\cdot)$ denotes the cumulative distribution function of the generalized Gaussian model.

The network is trained end-to-end by minimizing the rate-distortion cost:
\begin{equation}
\label{eq:rd}
L = \mathcal{R}(\hat{y}) + \mathcal{R}(\hat{z}) + \lambda \cdot \mathcal{D}(x, \hat{x})
\end{equation}
where $\mathcal{D}(x, \hat{x})$ measures the reconstruction distortion, and $\lambda$ controls the rate-distortion trade-off.

\begin{table*}[!t]
\centering
\addtolength{\abovedisplayskip}{20pt}   
\addtolength{\belowdisplayskip}{20pt}   
\renewcommand{\arraystretch}{1.2}   
\fontsize{8.4pt}{10.5}\selectfont
\setlength{\tabcolsep}{1.8pt}
\caption{Model configurations for different parameter scales.
The Base model corresponds to the original HPCM-Base\cite{li2025hpcm}, while the other configurations are scaled variants from 0.12B to 1B parameters for scaling law experiments. The definitions of $[L_1 \sim L_6]$, $[N_1 \sim N_3]$, and $[C_1 \sim C_6]$ are illustrated in Fig.~\ref{arch}. $Conv_{1\times 1}$ denotes a convolution layer with h kernel size as 1 and stride as 1.}
\vspace{0.8em}
\begin{tabular}{c|cc|cc|c|c|c}
\hline
\multirow{2}{*}{Model} & \multicolumn{2}{c|}{$g_a$/$g_s$}                                    & \multicolumn{2}{c|}{$h_a$/$h_s$}        & $g_{ep}^{S1}/g_{ep}^{S2}$   & $g_{ep}^{S3}$                  & \multirow{2}{*}{Params (M)} \\ \cline{2-7}
                       & {[}$L_1$, $L_2$, $L_3$, $L_4${]} & {[}$C_1$, $C_2$, $C_3$, $C_4${]} & {[}$L_5$, $L_6${]} & {[}$C_5$, $C_6${]} & {[}$N_1$, $N_2$, $N_3${]}      & {[}$N_1$, $N_2$, $N_3${]}      &                             \\ \hline
Base                   & {[}2, 2, 4, /{]}                 & {[}96, 192, 384, 320{]}          & {[}1, 3{]}         & {[}256, 256{]}     & {[}$Conv_{1\times 1}$, 2, 1{]} & {[}$Conv_{1\times 1}$, 3, 2{]} & 68.50                       \\
0.12B                  & {[}3, 3, 8, /{]}                 & {[}96, 192, 384, 320{]}          & {[}2, 4{]}         & {[}256, 256{]}     & {[}1, 2, 2{]}                  & {[}1, 3, 3{]}                  & 120.08                      \\
0.25B                  & {[}3, 3, 12, 3{]}                & {[}96, 192, 384, 320{]}          & {[}2, 4{]}         & {[}256, 256{]}     & {[}3, 3, 3{]}                  & {[}4, 4, 4{]}                  & 246.43                      \\
0.5B                   & {[}2, 2, 6, 2{]}                 & {[}192, 384, 768, 512{]}         & {[}4, 9{]}         & {[}512, 512{]}     & {[}2, 2, 2{]}                  & {[}3, 3, 3{]}                  & 543.57                      \\
1B                     & {[}3, 3, 9, 3{]}                 & {[}192, 384, 768, 512{]}         & {[}4, 9{]}         & {[}512, 512{]}     & {[}5, 5, 5{]}                  & {[}5, 6, 6{]}                  & 1002.00                     \\ \hline
\end{tabular}
\label{model_setting}
\end{table*}

\subsection{Model Scaling Strategy}
\label{scaling_arch}

For a better trade-off between compression performance and complexity, the previous HPCM-Base and HPCM-Large models \cite{li2025hpcm} contain only 68.5M and 89.7M parameters, respectively, constraining the representation capacity. To investigate large-scale learned image compression and the potential scaling laws, we progressively scale the parameter count of the HPCM-Base architecture shown in Fig.~\ref{arch}, as summarized in Table~\ref{model_setting}.
We fix the layer arrangement of the analysis/synthesis and hyper analysis/synthesis transforms and the HPCM entropy networks to enable a controlled comparison across sizes. The model parameters grow only through depth and width.
First, we increase the ResBlock counts [$L_1$, $L_2$, $L_3$, $L_4$, $L_5$, $L_6$] in the transform networks $g_a$/$g_s$/$h_a$/$h_s$, and the DepthConv block counts [$N_1$, $N_2$, $N_3$] in the entropy-parameter networks $g_{ep}^{S1}$/$g_{ep}^{S2}$/$g_{ep}^{S3}$. For the 0.5B and 1B models, we further widen the channel count[$C_1$, $C_2$, $C_3$, $C_4$, $C_5$, $C_6$] to provide higher capacity. 
The resulting models contain 68.50 M, 120.08 M, 246.43 M, 543.57 M, and 1002.00 M parameters, respectively.
Additionally, for the 1B model, we remove the cross attention-based context fusion module for stable training, using $\psi_{i-1}$ as the progressed context\cite{li2025hpcm}.
These models enable further scaling analysis in learned image compression.

\subsection{Scaling Analysis of Large Learned Image Compression Models}

\subsubsection{Background}
\label{scaling_backg}

Scaling laws describe how test loss changes as a function of one limiting resource (e.g., model size $N$, dataset size $D$, or optimal training compute $C_{min}$) when the others are fixed. Prior work on large language models shows that the loss $L$ typically follows a power law \cite{kaplan2020scaling, klug2022scaling, henighan2020scaling, rosenfeld2021scaling}:
\begin{equation}
\label{eq:scaling_law}
    L(X)=L_\infty+A X^{-\alpha}, \quad A>0,\ \alpha>0
\end{equation}
where $X$ can be any of $N$, $D$, or $C_{min}$.
This expression naturally separates into two components: an irreducible term $L_\infty$ and a reducible term $A X^{-\alpha}$. 
To interpret these two parts, we relate Eq.~\eqref{eq:scaling_law} to its probabilistic modeling form.
In many settings, including lossy image compression, the test loss includes the expected negative log-likelihood (NLL) under the true data distribution, which can be expressed as:
\begin{equation}
    \mathbb{E}_{x \sim P_{\text{true}}}\!\left[-\log P_{\text{model}}\right]
    = H\!\left(P_{\text{true}}\right) + D_{\mathrm{KL}}\!\left(P_{\text{true}}\,\|\,P_{\text{model}}\right),
\end{equation}
where $P_{\text{true}}$ and $P_{\text{model}}$ denote the true data distribution and the model distribution, respectively.
The $\mathbb{E}_{x \sim P_{\text{true}}}\!\left[-\log P_{\text{model}}\right]$ comprises two parts: the irreducible entropy $H(P_{\text{true}})$ of the data distribution, and the reducible divergence $D_{\text{KL}}$, which measures the gap between the model and the true distribution.
In Eq.~\eqref{eq:scaling_law}, the irreducible term $L_\infty$ corresponds to the entropy term $H(P_{\text{true}})$, while the reducible term $A X^{-\alpha}$ corresponds to the remaining modeling error. As $X \to \infty$, the reducible $A X^{-\alpha}$ vanishes and the loss approaches $L_\infty$. This implies that an infinitely large model could match the data distribution exactly.

In practice, following \cite{henighan2020scaling, tian2024visual}, we fit a single-term power law to the test loss $L$ as a function of $X$:
\begin{equation}
    L(X) = \gamma\, X^{-\alpha}, \quad \gamma>0,\ \alpha>0
\end{equation}
This can be viewed as fitting the reducible component after normalizing out the irreducible floor in Eq.~\eqref{eq:scaling_law}. We further take logarithms, and the relation becomes linear:
\begin{equation}
    \log L = -\alpha \log X + \log \gamma .
\end{equation}
We estimate $\alpha$ and $\gamma$ by ordinary least squares on the log-log scale.

\begin{figure*}[!t]
    \centering
    \includegraphics[width=1.0\linewidth]{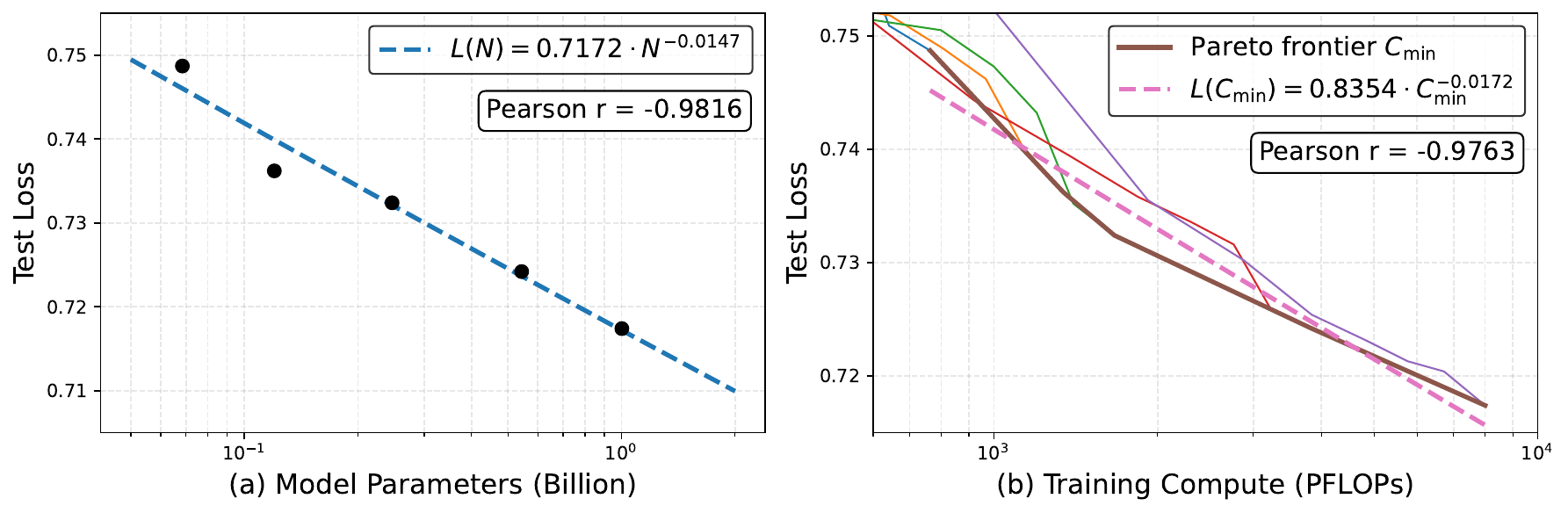}
    \caption{Scaling laws with (a) model parameters $N$ and (b) training compute $C$. All axes are plotted on a logarithmic scale. 
    In (a), black dots indicate five models with different sizes, and the blue dashed line shows the fitted power-law relation. 
    In (b), thin solid lines in different colors represent models with different sizes, while the thick brown solid line denotes the Pareto frontier of $L$, highlighting the optimal training compute $C_{\min}$ required to achieve a certain loss level. 
    The pink dashed line shows the corresponding power-law fit. 
    The small exponents $\alpha$ indicate a gradual decline in $L$ with increasing scale. The Pearson correlation coefficients near -0.98 reveal a relatively strong linear relationship between $\log N$ \textit{vs.} $\log L$ and $\log C_{\min}$ \textit{vs.} $\log L$.
    }
    \label{scaling_law}
\end{figure*}

\subsubsection{Scaling Laws with Model Size and Training Compute}

We use the 5 models with 68.5M to 1.0B parameters described in Sec.~\ref{scaling_arch}. We use the rate-distortion loss $L = \mathcal{R} + \lambda \cdot \mathcal{D}$ with $\lambda=0.013$ and mean-squared error $\mathcal{D}$ to validate the scaling law. The losses are tested on the Kodak\cite{franzen1999kodak} dataset.

\textbf{Scaling laws with model parameters $N$.} We first investigate how the test loss varies with model size. Following the methodology in Sec.\ref{scaling_backg}, we fit the linear relation in the log-log scale using SciPy’s \texttt{linregress} function.
As shown in Fig.~\ref{scaling_law} (a), the results reveal a clear power-law trend, particularly for the three largest models. The fitted scaling relation is:
\begin{equation}
    L(N)=0.7172 \cdot N^{-0.0147}
\end{equation}
The Pearson correlation coefficient of $r=-0.9816$ confirms a relatively strong linear relationship between $\log N$ and $\log L$. The small exponents $\alpha$ indicate a gradual decline in $L$ with increasing scale. These results confirm that scaling up the HPCM models consistently improves performance.

Using the fitted law, we can forecast the performance of larger scale models:
\begin{equation}
    L(2B) \approx 0.7099, \quad L(10B) \approx 0.6933
\end{equation}
When converted to BD-Rate, the 10B model corresponds to an approximate 30\% bitrate reduction at this rate point compared to VTM, highlighting the substantial potential gains achievable through continued scaling.

\textbf{Scaling laws with optimal training compute $C_{min}$.} We further investigate how test loss scales with optimal training compute. For each of the 5 model sizes, we track the test loss $L$ as a function of the training compute $C$ during training, measured in PFLOPs ($10^{15}$ floating point operations). As shown in Fig.~\ref{scaling_law}(b), we extract the Pareto frontier of $L$ to identify the optimal training compute $C_{min}$ required to reach a certain loss value. The fitted power-law relation is:
\begin{equation}
    L(C_{min})=0.8354 \cdot C_{min}^{-0.0172}
\end{equation}
This scaling trend indicates that when trained on sufficient data, larger models are more compute-efficient, reaching the same performance with fewer training FLOPs.

\begin{figure*}[!t]
    \centering
    \centerline{
        \begin{minipage}{0.5\linewidth}
         \includegraphics[width=\textwidth]{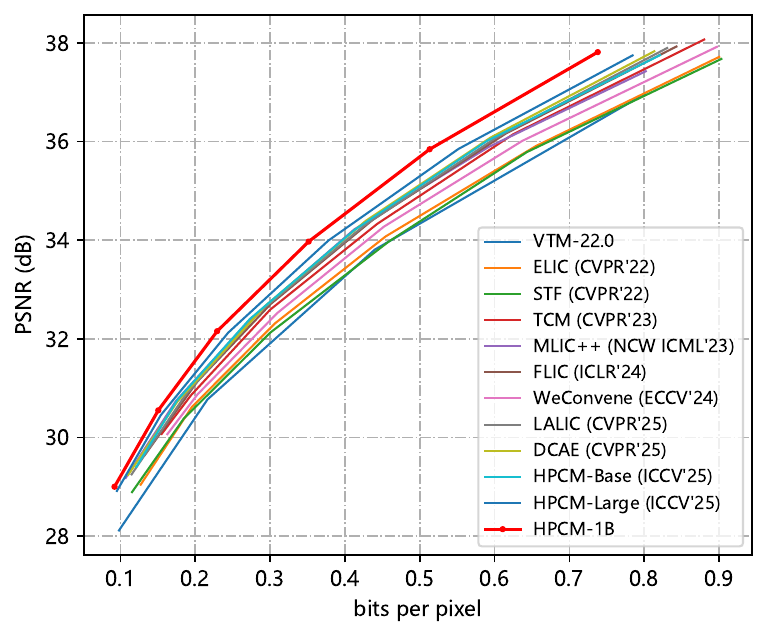}
        \end{minipage}
        \begin{minipage}{0.5\linewidth}
         \includegraphics[width=\textwidth]{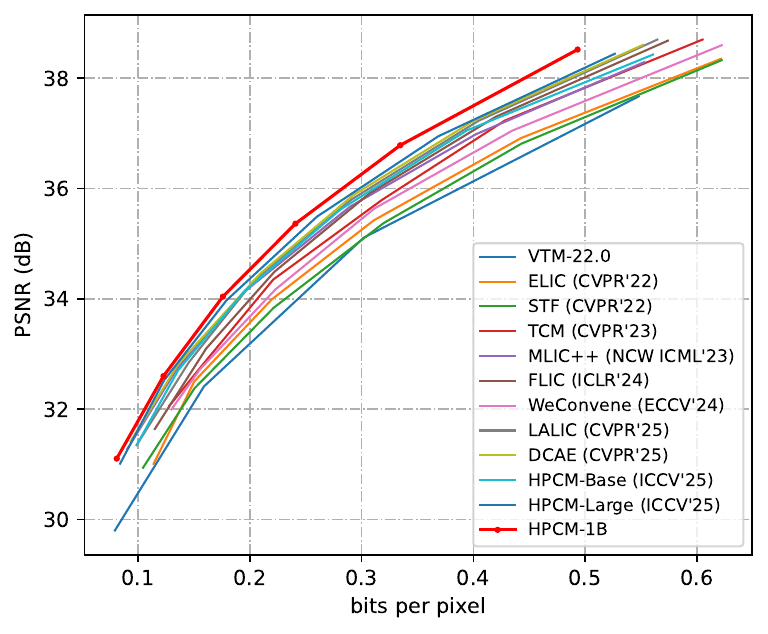}
        \end{minipage}
    }
    \caption{Rate-distortion curves on Kodak dataset (left) and Tecnick dataset (right).}
    \label{rd_curve}
\end{figure*}

\section{Experimental Results}

\subsection{Experimental Settings}

\setcounter{footnote}{0}
Our model implementation is based on the HPCM codebase\footnote{\url{https://github.com/lyq133/LIC-HPCM}}.

\textbf{Training settings.} All models are trained on the Flickr2W dataset~\cite{liu2020unified}. During training, images are randomly cropped to 256 × 256 patches with a batch size of 32. The optimization follows the rate-distortion objective in Eq.(\ref{eq:rd}), where distortion is measured using the mean squared error (MSE).
For the HPCM-1B model, we train at six different Lagrange multipliers $\lambda \in \{0.0018, 0.0035, 0.0067, 0.0130, 0.0250, 0.0483\}$ to produce a complete rate-distortion curve.
For the other scales (120.08 M, 246.43 M, 543.57 M), only a single rate point $\lambda =0.013$ is trained for scaling law fitting.
We adopt the Adam optimizer\cite{kingma2014adam} with $\beta_{1} = 0.9$ and $\beta_{2} = 0.999$. Our models are trained with 2 million training steps. The learning rate starts from $10^{-4}$, and reduced to $2\times 10^{-5}$ after 1.6M steps, then to $5\times 10^{-6}$ after 1.8M steps, and then to $10^{-6}$ after 1.9M steps.

\textbf{Evaluation settings.} We evaluate the compression performance on three commonly used test datasets: Kodak dataset \cite{franzen1999kodak} which contains 24 images with 512 × 768 resolution; CLIC Professional Validation (CLIC Pro Valid) dataset \footnote{\url{http://compression.cc}} which contains 41 high-quality images; Tecnick dataset \cite{asuni2014testimages} which contains 100 images with 1200 × 1200 resolution. 
Bitrate is measured in bits per pixel (bpp), and distortion is measured in peak signal-to-noise ratio (PSNR).
Rate savings are quantified using the BD-Rate metric~\cite{bjontegaard2001calculation}, with VTM-22.0\footnote{\url{https://vcgit.hhi.fraunhofer.de/jvet/VVCSoftware_VTM}} serving as the anchor.
Encoding and decoding times are measured on a single-core Intel(R) Xeon(R) Gold 6248R CPU and an NVIDIA GeForce RTX 3090 GPU.
Model complexity, including kMACs/pixel and parameter counts, is computed using the DeepSpeed library\footnote{\url{https://github.com/microsoft/DeepSpeed}}.

\begin{table*}[!t]
\centering
\setlength{\tabcolsep}{2.5pt}
\fontsize{8.5pt}{11}\selectfont
\caption{Compression performance and complexity comparison. VTM-22.0 is used as an anchor to calculate the PSNR BD-Rate. The best compression performance is marked in \textbf{bold}.}
\vspace{0.8em}
\begin{threeparttable}
\begin{tabular}{lccccccc}
\hline
\multirow{2}{*}{Model} & \multirow{2}{*}{\begin{tabular}[c]{@{}c@{}}Enc. Time$^\dagger$ \\ (ms)\end{tabular}} & \multirow{2}{*}{\begin{tabular}[c]{@{}c@{}}Dec. Time$^\dagger$ \\ (ms)\end{tabular}} & \multirow{2}{*}{\begin{tabular}[c]{@{}c@{}}kMACs\\ /pixel\end{tabular}} & \multirow{2}{*}{\begin{tabular}[c]{@{}c@{}}Params\\ (M)\end{tabular}} & \multicolumn{3}{c}{BD-Rate}                               \\ \cline{6-8} 
                       &                                                                            &                                                                            &                                                                         &                                                                       & Kodak             & CLIC Pro Valid    & Tecnick           \\ \hline
ELIC (CVPR'22) \cite{he2022elic}        & 126.5                                                                      & 111.4                                                                      & 573.88                                                                  & 36.93                                                                 & -3.22\%           & -3.89\%           & -4.57\%           \\
STF (CVPR'22) \cite{zou2022devil}         & 142.5                                                                      & 156.8                                                                      & 511.17                                                                  & 99.86                                                                 & -2.06\%           & 1.12\%            & -2.17\%           \\
TCM (CVPR'23) \cite{liu2023learned}         & 200.2                                                                      & 201.8                                                                      & 1823.58                                                                 & 76.57                                                                 & -10.70\%          & -8.32\%           & -11.84\%          \\
MLIC++ (NCW ICML'23) \cite{jiang2023mlic}  & 193.4                                                                      & 226.4                                                                      & 1282.81                                                                 & 116.72                                                                & -15.15\%          & -14.05\%          & -17.90\%          \\
FLIC (ICLR'24) \cite{li2023frequency}        & \textgreater{}1000                                                         & \textgreater{}1000                                                         & 1096.04                                                                 & 70.96                                                                 & -13.20\%          & -9.88\%           & -15.27\%          \\
WeConvene (ECCV’24) \cite{fu2024weconv}   & 343.6                                                                      & 256.5                                                                      & 2343.13                                                                 & 107.15                                                                & -6.98\%           & -5.66\%           & -8.63\%           \\
LALIC (CVPR'25) \cite{Feng_2025_CVPR}       & 189.0                                                                      & 95.4                                                                       & 667.26                                                                  & 66.13                                                                 & -14.09\%          & -14.22\%          & -18.31\%          \\
DCAE (CVPR'25) \cite{Lu_2025_CVPR}        & 134.6                                                                      & 132.4                                                                      & 940.40                                                                  & 119.4                                                                 & -15.36\%          & -15.40\%          & -20.35\%          \\
HPCM-Base (ICCV'25) \cite{li2025hpcm}   & 81.8                                                                       & 81.3                                                                       & 918.57                                                                  & 68.5                                                                  & -15.31\%          & -14.23\%          & -18.16\%          \\
HPCM-Large (ICCV'25) \cite{li2025hpcm}  & 91.2                                                                       & 90.2                                                                       & 1261.29                                                                 & 89.71                                                                 & -19.19\%          & -18.37\%          & -22.20\%          \\ \hline
HPCM-1B                & 350.9                                                                      & 342.5                                                                      & 9625.24                                                                 & 1002.00                                                               & \textbf{-24.21\%} & \textbf{-23.41\%} & \textbf{-25.68\%} \\ \hline
\end{tabular}
\begin{tablenotes}
\footnotesize
\item $^\dagger$ Coding time includes network inference time and arithmetic coding time. The arithmetic coding time varies across models due to different implementations in the released code and test environments.
\end{tablenotes}
\end{threeparttable}
\label{rd_time}
\end{table*}

\begin{figure*}[!t]
    \centering
    \centerline{
        \begin{minipage}{0.48\linewidth}
         \includegraphics[width=\textwidth]{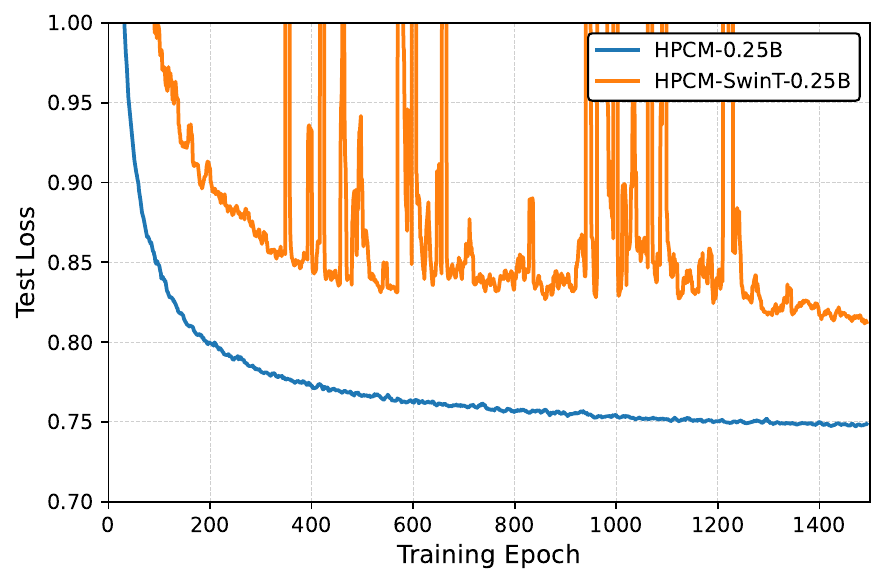}
        \end{minipage}
        \hspace{0.4cm}
        \begin{minipage}{0.48\linewidth}
         \includegraphics[width=\textwidth]{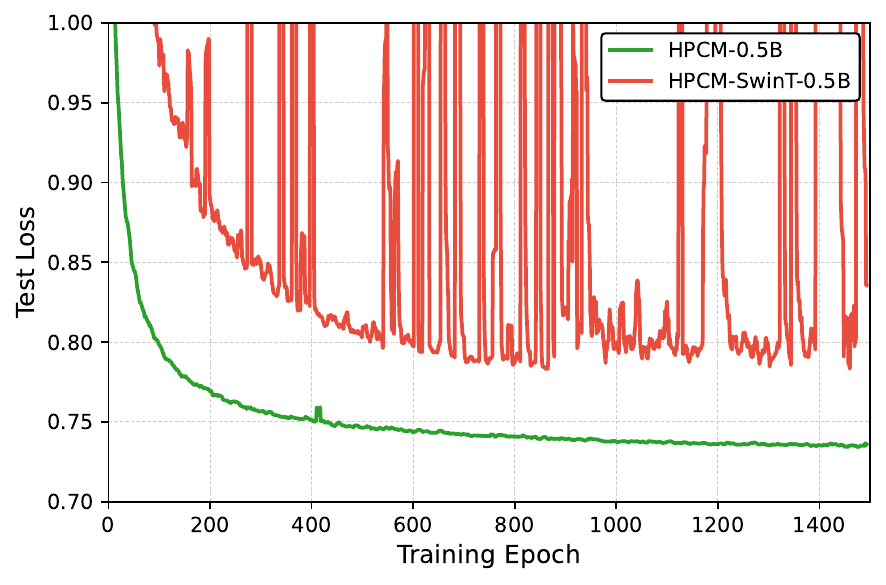}
        \end{minipage}
    }
    \caption{Test loss curves on the Kodak dataset for scaled Swin Transformer-based transform networks ($g_a$ and $g_s$) at 0.25B (left) and 0.5B (right) parameters, compared with the original convolution-based HPCM models of the same sizes. We visualize the test loss over the first 1500 training epochs.}
    \label{swin_test_loss}
\end{figure*}

\subsection{Rate-Distortion Performance and Complexity}

We primarily compare our 1 billion parameter model, HPCM-1B, to state-of-the-art (SOTA) learned image compression approaches\cite{he2022elic, zou2022devil, liu2023learned, jiang2023mlic, li2023frequency, fu2024weconv, Feng_2025_CVPR, Lu_2025_CVPR, li2025hpcm}. 
As shown in Fig.\ref{rd_curve}, HPCM-1B consistently outperforms other advanced methods in terms of PSNR across the entire bitrate range. On both the Kodak and Tecnick datasets, it achieves up to $\sim$0.3dB higher PSNR at high bitrates compared with the best existing models.
Table \ref{rd_time} shows the BD-Rate performance of various methods. Compared to VTM-22.0, our HPCM-1B model achieves 24.21\%, 23.41\%, and 25.68\% bitrate savings on Kodak, CLIC Pro Valid, and Tecnick datasets, respectively.

Table~\ref{rd_time} also reports the model complexity, including encoding/decoding time, kMACs per pixel, and parameter count.
Due to its larger scale, HPCM-1B naturally incurs a higher computational cost.

\subsection{Scaling Transformer-based Architectures}

Transformers~\cite{vaswani2017attention, dosovitskiy2020image} have demonstrated remarkable capability in large-scale models across various domains. In this work, we also explore scaling Swin Transformer-based~\cite{liu2021swin, zhu2022transformer} transform networks ($g_a$ and $g_s$) in learned image compression, while keeping the HPCM entropy model unchanged. We scale the model size to around 0.25B and 0.5B parameters, and compare them with the original HPCM models of the same sizes. 

Figure~\ref{swin_test_loss} plots the test loss on the Kodak dataset over training epochs. The Swin Transformer-based models exhibit unstable convergence and higher test loss compared to the original convolution-based HPCM variants. 
These results suggest that, although Transformers are widely regarded as powerful architectures for building large-scale models, their application to large-scale learned image compression remains underexplored.

\section{Limitations}

While our study provides valuable insights into the scaling behavior of large learned image compression models, several limitations remain.

\noindent\textbf{(1) Model generality.} Our scaling-up experiments primarily focus on enlarging the parameter count of the HPCM architecture and its Transformer-based variant. We have not explored scaling trends in other model designs, which may exhibit distinct behaviors.

\noindent\textbf{(2) Model scale granularity.} Due to computational constraints, the scaling law curves are fitted using only five model sizes. Incorporating a larger set of model configurations with varying parameter counts would yield more robust and reliable scaling law estimates.

\noindent\textbf{(3) Training strategies.} We have not explored advanced training strategies for large-scale learned image compression, such as leveraging more diverse or higher-quality datasets, improved learning rate schedules, or other optimization techniques. Combining such strategies with greater computational budgets could further unlock the performance potential of large-scale models.

\section{Conclusion}

In this work, we present the first study on scaling up learned image compression models and characterizing their performance through scaling laws. Building upon the HPCM architecture, we train models ranging from 68.5M to 1B parameters and fit power-law relationships between test loss and key scaling variables. Our analysis reveals consistent scaling trends, enabling further extrapolation to even larger models. Experimental results also show that the scaled-up HPCM-1B achieves state-of-the-art rate-distortion performance. We hope this work provides both a practical benchmark and a conceptual foundation for future research on large-scale learned compression models and deeper exploration of the link between compression and intelligence.

\bibliographystyle{plain}  
\small
\bibliography{main}





\end{document}